\documentclass[runningheads,a4paper]{llncs}
\usepackage{amsmath}
\usepackage{amssymb}
\usepackage{graphicx}
\usepackage{booktabs}
\usepackage{listings}
\usepackage{xcolor}
\usepackage{hyperref}
\usepackage{caption}
\usepackage{subcaption}

\usepackage{url}
\usepackage{color}
\usepackage{cite}
\usepackage{epsfig}

\usepackage[nomargin,inline,marginclue,draft]{fixme}
\usepackage{cleveref}
\fxsetup{status=draft}
\fxsetup{theme=color, mode=multiuser} \FXRegisterAuthor{me}{ame}{\color{red}Me}
\newcommand{\orcid}[1]{\href{https://orcid.org/#1}{\includegraphics[scale=0.02]{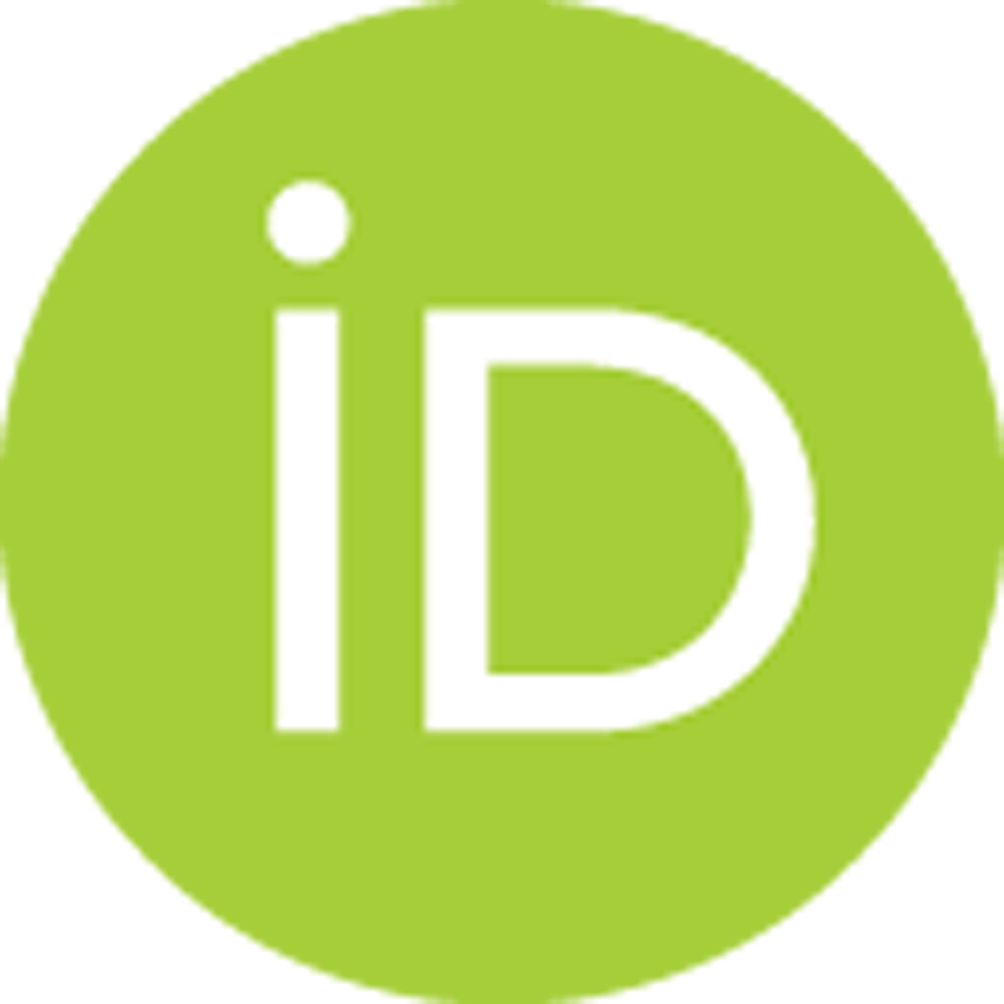}}} 

\hypersetup{
    colorlinks=true
}
\title{The RSNA-ASNR-MICCAI BraTS 2021 Benchmark on Brain Tumor Segmentation and Radiogenomic Classification}
\titlerunning{The RSNA-ASNR-MICCAI BraTS 2021 Benchmark}
\author{
Ujjwal Baid\inst{1,2,3,\dag,\ddag,\orcid{0000-0001-5246-2088}}
\and Satyam Ghodasara\inst{2,\dag,\orcid{0000-0002-5332-3132}}
\and Suyash Mohan\inst{1,2,\dag,\orcid{0000-0002-4025-115X}}
\and Michel Bilello\inst{1,2,\dag}
\and Evan Calabrese\inst{4,\dag,\ddag,\S,\orcid{0000-0002-1464-0354}}
\and
Errol Colak\inst{5,\dag,\ddag}
\and Keyvan Farahani\inst{6,\dag,\orcid{0000-0003-2111-1896}}
\and Jayashree Kalpathy-Cramer\inst{7,\dag}
\and Felipe C. Kitamura\inst{9,10,\dag,\orcid{0000-0002-9992-5630}}
\and Sarthak Pati\inst{1,2,3,11,\dag,\orcid{0000-0003-2243-8487}}
% ...
}

\authorrunning{Baid et al.}

\institute{\scriptsize{Center for Biomedical Image Computing and Analytics (CBICA), University of Pennsylvania, Philadelphia, PA, USA
\and
Department of Radiology, Perelman School of Medicine at the University of Pennsylvania, Philadelphia, PA, USA
\and
Department of Pathology and Laboratory Medicine, Perelman School of Medicine at the University of Pennsylvania, Philadelphia, PA, USA
\and
Department of Radiology \& Biomedical Imaging, University of California San Francisco, CA, USA
\and
Unity Health Toronto, University of Toronto, Toronto, ON 
\and
Center for Biomedical Informatics and Information Technology, National Cancer Institute, National Institutes of Health
\and
Athinoula A. Martinos Center for Biomedical Imaging, MGH/Harvard Medical School, MA, USA
\and
Department of Radiology, Thomas Jefferson University, Philadelphia, PA, USA
\and
Universidade Federal de São Paulo (UNIFESP), São Paulo, Brazil
\and
Diagnósticos da América SA (DASA), São Paulo, Brazil
\and
Department of Informatics, Technical University of Munich, Munich, Germany
\and
The Ohio State University Wexner Medical Center, Columbus, OH, USA 
\and
Penn Statistics in Imaging and Visualization Center, Department of Biostatistics, Epidemiology and Informatics, Perelman School of Medicine, University of Pennsylvania, Philadelphia, PA, USA 
\and
Sage Bionetworks, Seattle, WA, USA
\and
Kaggle
\and
Center for Intelligent Imaging, University of California San Francisco, CA, USA
\and
Leidos Biomedical Research, Inc., Frederick National Laboratory for Cancer Research, Frederick, MD 21701, USA 
\and
Department of Neuroradiology, Klinikum rechts der Isar, School of Medicine, Technical University of Munich, München, Germany 
\and
Department of Diagnostic Radiology, University of Texas MD Anderson Cancer Center, Houston, TX, USA
\and
Hillman Cancer Center, University of Pittsburgh Medical Center, Pittsburgh, PA, 15232, USA
\and
Neuroimaging Informatics and Analysis Center, Washington University in St. Louis, St. Louis, MO, USA
\and
Department of Radiology, Washington University in St. Louis, St. Louis, MO, USA
\and
Department of Neurology, The University of Alabama at Birmingham, Birmingham, AL, USA
\and
Institute for Surgical Technology and Biomechanics, University of Bern, Bern, Switzerland
\and
Support Centre for Advanced Neuroimaging Inselspital, Institute for Diagnostic and Interventional Neuroradiology,Bern University Hospital, Bern, Switzerland
\and
Center for MR-Research, University Children’s Hospital Zurich, Zurich, Switzerland
\and
Institute of Diagnostic and Interventional Radiology, Pediatric Radiology and Neuroradiology, University Medical Center Rostock, Rostock, Germany
\and
Department of Radiodiagnosis, Tata Memorial Hospital, Tata Memorial Centre, Homi Bhabha National Institute, Mumbai, India
\and
Department of Quantitative Biomedicine, University of Zurich, Zurich, Switzerland
\and
Mercy catholic medical center, Darby, PA, USA
\and
Diagnóstico Avançado por Imagem and Hospital Erasto Gaertner, Curtiba, Brazil
\and
Department of Medical Imaging, Gold Coast University Hospital, Southport, Australia
\and
Department of Radiology, University of Ottawa, Ottawa, Canada
\and
Department of Radiology, Baylor College of Medicine, Houston, TX, USA
\and
Department of Radiology, Hennepin Healthcare, Minneapolis, MN, USA 
\and
Uniformed Services University, Bethesda, MD, USA
\and
Institute of Neurology, University College London, London, United Kingdom
\and
Virtual Radiologic Professionals, LLC - Branson, Eden Prairie, MN, USA
\and
University of Pittsburgh Medical Center, Pittsburg, PA, USA
\and
Hahnemann University Hospital Drexel University College of Medicine, PA, USA
\and
Department of Radiology, Oregon Health \& Science University, Portland, OR, USA
\and
Dr Jones and Partners Medical Imaging, South Australia
\and
Department of Neuroradiology. Hospital Italiano de Buenos Aires, Buenos Aires, Argentina
\and
Mallinckrodt Institute of Radiology, Washington University in St. Louis, MO, USA
\and
Neuroradiology Department, Hospital Garcia de Orta EPE, Almada, Portugal
\and
Department of Radiology, Tufts MedicalCenter, Boston, MA, USA.
\and
Centro Universitario Euro-Americana (UNIEURO), Brasília, DF, Brazil
\and
Department of Radiology, Division of Neuroradiology, University of Southern California, Keck School of Medicine, Los Angeles, CA, USA.
\and
Radiology (Neuroradiology Section), University of Louisville, Louisville, KY, USA
\and
Northwestern University Feinberg School of Medicine, Chicago, IL, USA
\and
 Stanford Hospital and Clinics, Stanford University, Stanford, CA, USA
\and
Department of Radiology and Medical Imaging University of Virginia Health System Charlottesville, VA, USA
\and
Department of Radiology, Baylor College of Medicine, Houston, Tex, USA
\and
NYU Langone Medical Center, New York, NY, USA
\and
Kaiser Permanente, San Diego, CA, USA
\and
Instituto Nacional de Ciencias Medicas y Nutricion, Maxico City, Maxico
\and
Northport VA Medical Center Northport, NY, USA
\and
Ohio State University Wexner Medical Center, Columbus, OH, USA
\and
University of Virginia Medical Center, Charlottesville, VA, USA
\and
Department of Radiology, Kingston General Hospital, Queen's University, Kingston, Canada
\and
Department of Radiology, University of Minnesota Health,Minneapolis, MN, USA
\and
NorthShore University HealthSystem, Chicago, IL, USA
\and
Neuroradiology and Imaging Informatics Imaging Institute, Cleveland Clinic, Cleveland, OH, USA
\and
University of Utah Health Sciences Center, Salt Lake City, UT, USA
\and
London Health Sciences Centre, London, Ontario, Canada
\and
Dr Abdurrahman Yurtaslan Ankara Oncology Training and Research Hospital, University of Health Sciences, Ankara, Turkey
\and
University of Texas Health San Antonio, TX, USA
\and
Department of Radiology, Beth Israel Deaconess Medical Center, Harvard Medical School, Boston, MA, USA
\and
Neuroradiology and Imaging Informatics Imaging Institute, Cleveland Clinic, Cleveland, OH, USA
\and
Emory University, Atlanta, GA, USA
\and
Memorial Sloan Kettering Cancer Center, New York, NY, USA
\and
Yale University School of Medicine, Department of Radiology \& Biomedical Imaging, New Haven, CT, USA
\and
Mayo Clinic, Jacksonville, FL, USA
\and
Dartmouth Hitchcock Medical Center, NH, USA
\and
Oakland University William Beaumont School of Medicine, Rochester, MI, USA
\and
 Radiology Department at Klinikum Hochrhein Waldshut-Tiengen, Germany
\and
Maastricht University Hospital, Maastricht, The Netherlands
\and
Radiology department,Centre Hospitalier de l'Universite de Montreal (CHUM), Centre de Recherche du CHUM (CRCHUM) Montreal, Quebec, Canada
\and
Department of Radiology, NewYork-Presbyterian Hospital, Weill Cornell Medical College, New York, NY, USA
\and
University of Iowa Hospitals and Clinics, Iowa City, IA, USA
\and
Department of Radiology Stanford University, Stanford, CA, USA
\and
Department of Radiology, Northwell Health, Zucker Hofstra School of Medicine at Northwell, North Shore University Hospital, Hempstead, New York, NY, USA.
\and
Department of Medical Imaging, University of Toronto, ON, Canada
\and
University of California Los Angeles, CA, USA
\and
University of Rochester Medical Center,Rochester, NY, USA
\and
Lahey Clinic, Burlington, MA, USA
\and
Temple University Hospital, Philadelphia, PA, USA
}
\\
\textsuperscript{\dag} People involved in the organization of the challenge.\\
\textsuperscript{\ddag} People contributing data from their institutions.\\
\textsuperscript{\S} People involved in annotation process.\\ 
\textsuperscript{\P} Equal senior authors.\\ 
\textsuperscript{*} Corresponding author: \email{\{sbakas@upenn.edu\}}}

\begin{document}
\mainmatter
\maketitle
\setcounter{footnote}{0} 
\begin{abstract}
    The BraTS 2021 challenge celebrates its $10^{th}$ anniversary and is jointly organized by the Radiological Society of North America (RSNA), the American Society of Neuroradiology (ASNR), and the Medical Image Computing and Computer Assisted Interventions (MICCAI) society. Since its inception, BraTS has been focusing on being a common benchmarking venue for brain glioma segmentation algorithms, with well-curated multi-institutional multi-parametric Magnetic Resonance Imaging (mpMRI) data. Gliomas are the most common primary malignancies of the central nervous system, with varying degrees of aggressiveness and prognosis. The RSNA-ASNR-MICCAI BraTS 2021 challenge targets the evaluation of computational algorithms assessing the same tumor compartmentalization, as well as the underlying tumor’s molecular characterization, in pre-operative baseline mpMRI data from 2,040 patients. Specifically, the two tasks that BraTS 2021 focuses on are: a) the segmentation of the histologically distinct brain tumor sub-regions, and b) the classification of the tumor’s O[6]-methylguanine-DNA methyltransferase (MGMT) promoter methylation status. The performance evaluation of all participating algorithms in BraTS 2021 will be conducted through the Sage Bionetworks Synapse platform (Task 1) and Kaggle (Task 2), concluding in distributing to the top ranked participants monetary awards of \$60,000 collectively.
\end{abstract}


\section{Introduction}

    Glioblastoma (GBM), and diffuse astrocytic glioma with molecular features of GBM (WHO Grade 4 astrocytoma), are the most common and aggressive malignant primary tumor of the central nervous system (CNS) in adults, with extreme intrinsic heterogeneity in appearance, shape, and histology \cite{louis2019cimpact,cimpact_1,cimpact_2,cimpact_3,cimpact_4,cimpact_5,cimpact_6}. GBM patients have an average prognosis of 14 months, following standard of care treatment (comprising surgical resection followed by radiotherapy and chemotherapy), and 4 months left untreated \cite{OS_SB}. Although various experimental treatment options have been proposed during the past 20 years, there have not been any substantial differences in patient prognosis.
     
    Accurate identification of brain tumor sub-regions boundaries in MRI is of profound importance in many clinical applications, such as surgical treatment planning, image-guided interventions, monitoring tumor growth, and the generation of radiotherapy maps. However, manual detection and tracing of tumor sub-regions is tedious, time-consuming, and subjective. In a clinical setup, this manual process is carried out by radiologists in a qualitative visual manner, and hence becomes impractical when dealing with numerous patients. This highlights the unmet need for automated deterministic segmentation solutions that could contribute in expediting this process.
     
    The release of the current revised World Health Organization (WHO) classification of CNS tumors \cite{WHO_louis20162016} highlighted the appreciation of integrated diagnostics, and transitioned the clinical tumor diagnosis from a purely morphologic-histopathologic classification to integrating molecular-cytogenetic characteristics. O[6]-methylguanine-DNA methyltransferase (MGMT) is a DNA repair enzyme that the methylation of its promoter in newly diagnosed GBM has been identified as a favorable prognostic factor and a predictor of chemotherapy response \cite{MGMT}. Thus, determination of MGMT promoter methylation status in newly diagnosed GBM can influence treatment decision making.
     
    The RSNA ASNR MICCAI Brain Tumor Segmentation (BraTS) 2021 challenge utilizes multi-institutional multi-parametric Magnetic Resonance Imaging (mpMRI) scans, to address both the automated tumor sub-region segmentation and the prediction of one of the genetic characteristics of glioblastoma (MGMT promoter methylation status) from pre-operative baseline MRI scans. Specifically, BraTS 2021 focuses on the evaluation of state-of-the-art methods for the accurate segmentation of intrinsically heterogeneous brain glioma sub-regions and on the evaluation of classification methods distinguishing between MGMT methylated (MGMT+) and unmethylated (MGMT-) tumors. This manuscript describes the characteristics of the data included in the BraTS 2021 challenge, along with the annotation protocol followed to prepare the challenge data, an elaborate description of the challenge’s tasks, and the performance evaluation of all participating methods (in Section 2) and then discusses the limitations and currently considered future directions (in Section 3).

\section{Materials \& Methods}
    \subsection{Data}
\label{sec:data}

    The BraTS dataset describes a retrospective collection of brain tumor mpMRI scans acquired from multiple different institutions under standard clinical conditions, but with different equipment and imaging protocols, resulting in a vastly heterogeneous image quality reflecting diverse clinical practice across different institutions. Inclusion criteria comprised pathologically confirmed diagnosis and available MGMT promoter methylation status. These data have been updated, since BraTS 2020 \cite{menze2014multimodal, bakas2017advancing, bakas2018identifying, bakas2017segmentation_1, bakas2017segmentation_2}, increasing the total number of cases from 660 to 2,000. Ground truth annotations of every tumor sub-region for task 1 were approved by expert neuroradiologists, whereas the MGMT methylation status was based on the laboratory assessment of the surgical brain tumor specimen \cite{calabrese2021university}.
 
    Following the paradigm of algorithmic evaluation in machine learning, the data included in the BraTS 2021 challenge are divided in training, validation, and testing datasets. The challenge participants are provided with the ground truth labels only for the training data. The validation data are then provided to the participants without any associated ground truth and the testing data are kept hidden from the participants at all times. 
     
    Participants are not allowed to use additional public and/or private data (from their own institutions) for extending the provided BraTS data, for the training of the algorithm chosen to be ranked. Similarly, using models that were pretrained on such datasets is not allowed. This is due to our intentions to provide a fair comparison among the participating methods. However, participants are allowed to use additional public and/or private data (from their own institutions), only for scientific publication purposes and if they explicitly mention this in their submitted manuscripts. Importantly, participants that decide to proceed with this scientific analysis they must also report results using only the BraTS'21 data to discuss potential result differences.
    
    \subsubsection{Imaging Data Description} 
    
    The mpMRI scans included in the BraTS 2021 challenge describe a) native (T1) and b) post-contrast T1-weighted (T1Gd (Gadolinium)), c) T2-weighted (T2), and d) T2 Fluid Attenuated Inversion Recovery (T2-FLAIR) volumes, acquired with different protocols and various scanners from multiple institutions. 
    
    Standardized pre-processing has been applied to all the BraTS mpMRI scans. Specifically, the applied pre-processing routines include conversion of the DICOM files to the NIfTI file format \cite{nifti}, co-registration to the same anatomical template (SRI24) \cite{SRI_rohlfing2010sri24}, resampling to a uniform isotropic resolution ($1mm^{3}$), and finally skull-stripping. The pre-processing pipeline is publicly available through the Cancer Imaging Phenomics Toolkit (CaPTk) \footnote{\url{https://cbica.github.io/CaPTk/}} \cite{captk,captk_2,captk_3} and Federated Tumor Segmentation (FeTS) tool \footnote{\url{https://github.com/FETS-AI/Front-End/}}. Conversion to NIfTI strips the accompanying metadata from the DICOM images, and essentially removes all Protected Health Information (PHI) from the DICOM headers. Furthermore, skull-stripping mitigates potential facial reconstruction/recognition of the patient \cite{NEJMc1908881,NEJMc1915674}. The specific approach we have used for skull stripping is based on a novel DL approach that accounts for the brain shape prior and is agnostic to the MRI sequence input \cite{thakur2020brain}.

\begin{figure}[t]
          \centering
          \includegraphics[width=1\linewidth]{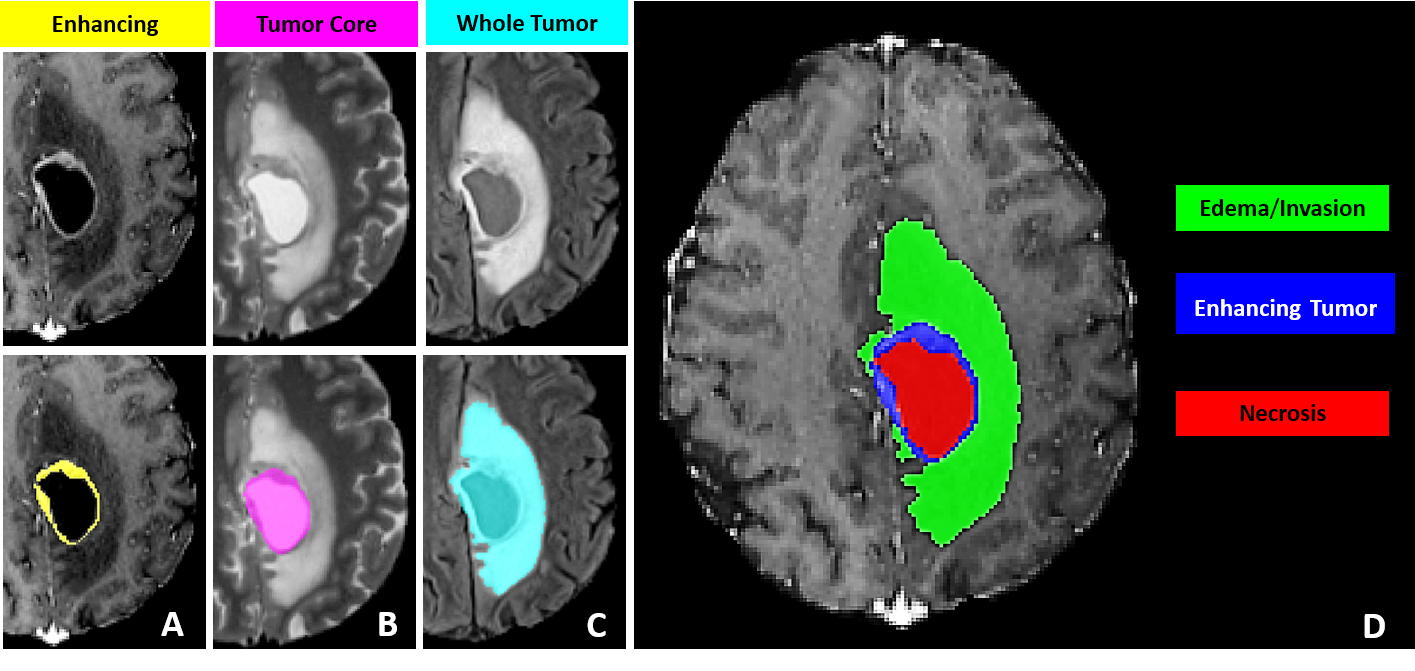}  
          \caption{\textbf{Glioma sub-regions considered in the RSNA-ASNR-MICCAI BraTS 2021 challenge.} Image panels with the tumor sub-regions annotated in the different mpMRI scans. The image panels A-C denote the regions considered for the performance evaluation of the participating algorithms and specifically highlight (from left to right): the enhancing tumor (ET - yellow) visible in a T1Gd scan, surrounding the cystic/necrotic components of the core (panel A), the tumor core (TC – magenta) and the whole tumor (WT - cyan) visible in the corresponding T2 (panel B) and T2-FLAIR (panel C) scans, respectively. Panel D, depicts the combined segmentations generating the final tumor sub-region labels, as provided to the BraTS 2021 participants: enhancing core (yellow), necrotic/cystic core (red), and edema/invasion (green).}
        \label{annotations}
    \end{figure}

    Specifically for Task 1 (Tumor sub-region segmentation), all imaging volumes have then been segmented using the STAPLE \cite{warfield2004simultaneous} fusion of previous top-ranked BraTS algorithms, namely, nnU-Net \cite{isensee2020nnu}, DeepScan \cite{mckinley2018ensembles}, and DeepMedic \cite{kamnitsas2017efficient}. These fused labels were then refined manually by volunteer neuroradiology experts of varying rank and experience, following a consistently communicated annotation protocol. The manually refined annotations were finally approved by experienced board-certified attending neuro-radiologists, with more than 15 years of experience working with gliomas. The annotated tumor sub-regions are based upon known observations visible to the trained radiologist (VASARI features) and comprise the Gd-enhancing tumor (ET — label 4), the peritumoral edematous/invaded tissue (ED — label 2), and the necrotic tumor core (NCR — label 1). ET is the enhancing portion of the tumor, described by areas with both visually avid, as well as faint, enhancement on T1Gd MRI. NCR is the necrotic core of the tumor, the appearance of which is hypointense on T1Gd MRI. ED is the peritumoral edematous and infiltrated tissue, defined by the abnormal hyperintense signal envelope on the T2 FLAIR volumes, which includes the infiltrative non enhancing tumor, as well as vasogenic edema in the peritumoral region. The tumor sub-regions are shown in Fig. \ref{annotations}. 
    
    For Task 2 (Radiogenomic Classification), all the imaging volumes were first pre-processed as for Task 1, in order to obtain the skull-stripped volumes, which are then converted from NIfTI to DICOM file format. To facilitate this conversion, while ensuring that the original patient space is preserved, the skull-stripped brain volumes in NIfTI format of each MRI sequence and their corresponding original DICOM scans (in the patient space) are required. The DICOM volume is read as an ITK image \cite{ITK} and the skull-stripped volume is rigidly registered to it, providing a transformation matrix that defines the spatial mapping between the 2 volumes. This transformation matrix is applied to the skull-stripped volume and to the corresponding segmentation labels, to translate them both to the patient space. This process is taking place for each modality separately. The transformed volumes are then passed through CaPTk's NIfTI to DICOM conversion engine to generate DICOM image volumes for the skull-stripped images. Once all mpMRI sequences were converted back to the DICOM file format, further de-identification took place based on a two-step process. The first step used the RSNA CTP (Clinical Trials Processor) Anonymizer  \footnote{\url{http://mirc.rsna.org/download/Anonymizer-installer.jar}} with the standard built-in script. Step two then consisted of whitelisting the DICOM files from step 1. The whitelisting process removes all non-essential tags from the DICOM header. This last process ensures there are no protected health information (PHI) entries left in the DICOM header.

    \subsubsection{MGMT Promoter Methylation Data Description}
        
        The MGMT promoter methylation status data is defined as a binary label (0: unmethylated, 1: methylated), and provided to the participants as a comma-separated value (.csv) file with the corresponding pseudo-identifiers of the mpMRI volumes (study-level label). 
 
        The MGMT promoter methylation status of the BraTS 2021 dataset was determined at each of the host institutions based on various techniques, including pyrosequencing, and next generation quantitative bisulfite sequencing of promoter CpG sites. Sufficient tumor tissue collected at time of surgery was required for both approaches. For the pyrosequencing approach, the genomic DNA was initially extracted from 5lm tissue sections of formalin-fixed paraffin-embedded (FFPE) tissue samples. DNA was further cleaned and purified. The DNA concentration, protein to nucleic acid ratio, and DNA to RNA ratio for purity were assessed by spectrophotometer. Approximately 500–1000ng total DNA was subjected to bisulfite conversion using the EPiTect Bisulfite Kit. A total of 50–100 ng bisulfite-treated DNA was carried on for PCR using F-primer and R-primer. Pyrosequencing methylation assay was then conducted using the sequencing primer on the PyroMark Q96ID pyrosequencer. The Pyromark CpG MGMT kit detected the average level of methylation on CpG 74–81 sites located in the MGMT gene. A cytosine not followed by a guanine served as an internal control for completion of bisulfite conversion. The percent methylation above 10\% was interpreted as positive. A sample below 10\% methylation was interpreted as negative. For the latter approach, a total of 17 MGMT promoter CpG sites were amplified by nested polymerase chain reaction (PCR) using a bisulfite treated DNA template. Quantitative PCR was performed for each CpG site to determine its methylation status. A result of 2\% or more methylated CpG sites in the MGMT promoter (out of 17 total sites) was considered a positive result.

    \subsubsection{Comparison with Previous BraTS datasets}
       
       \begin{table}[t]
       \caption{Summary of distribution of BraTS Challenge data across training, validation and test cohorts since the inception of BraTS initiative. (TBA: To Be Announced)}
       \label{Table1}
        \begin{tabular}{|l|l|l|l|l|l|l|}
        \hline
        \textbf{Year}                                              & \textbf{\begin{tabular}[c]{@{}l@{}}Total\\  Data\end{tabular}} & \textbf{\begin{tabular}[c]{@{}l@{}}Training\\ Data\end{tabular}} & \textbf{\begin{tabular}[c]{@{}l@{}}Validation\\ Data\end{tabular}} & \textbf{\begin{tabular}[c]{@{}l@{}}Testing\\ Data\end{tabular}} & \textbf{Tasks}                                                                    & \textbf{Timepoint} \\ \hline
        2012                                                       & 50                                                             & 35                                                               & NA                                                                 & 15                                                              & Segmentation                                                                      & Pre-operative  \\ \hline
        2013                                                       & 60                                                             & 35                                                               & NA                                                                 & 25                                                              & Segmentation                                                                      & Pre-operative  \\ \hline
        2014                                                       & 238                                                            & 200                                                              & NA                                                                 & 38                                                              & Segmentation                                                                      & Longitudinal       \\ \hline
        2015                                                       & 253                                                            & 200                                                              & NA                                                                 & 53                                                              & \begin{tabular}[c]{@{}l@{}}Segmentation\\ Disease progression\end{tabular}        & Longitudinal       \\ \hline
        2016                                                       & 391                                                            & 200                                                              & NA                                                                 & 191                                                             & \begin{tabular}[c]{@{}l@{}}Segmentation\\ Disease progression\end{tabular}        & Longitudinal       \\ \hline
        2017                                                       & 477                                                            & 285                                                              & 46                                                                 & 146                                                             & \begin{tabular}[c]{@{}l@{}}Segmentation\\ Survival prediction\end{tabular}        & Pre-operative  \\ \hline
        2018                                                       & 542                                                            & 285                                                              & 66                                                                 & 191                                                             & \begin{tabular}[c]{@{}l@{}}Segmentation\\ Survival prediction\end{tabular}        & Pre-operative \\ \hline
        2019                                                       & 626                                                            & 335                                                              & 125                                                                & 166                                                             & \begin{tabular}[c]{@{}l@{}}Segmentation\\ Survival prediction\end{tabular}        & Pre-operative  \\ \hline
        2020                                                       & 660                                                            & 369                                                              & 125                                                                & 166                                                             & \begin{tabular}[c]{@{}l@{}}Segmentation\\ Survival prediction\end{tabular}        & Pre-operative \\ \hline
        \begin{tabular}[c]{@{}l@{}}2021 \end{tabular} & 2040                                                           & 1251                                                             & 219                                                                & 570                                                             & \begin{tabular}[c]{@{}l@{}}Segmentation\\ MGMT classification\end{tabular} & Pre-operative \\ \hline
        \end{tabular}
        \end{table}

        The first BraTS challenge was organized in 2012 in conjunction with the MICCAI conference, and was making available a total of 50 mpMRI glioma cases (Table \ref{Table1}). The BraTS'12-'13 dataset was manually annotated by clinical experts, and the task at hand was the segmentation of the glioma sub-regions (ET, NCR, ED). In BraTS'14-'16 the dataset provided to the participants included a large contribution of data from The Cancer Imaging Archive (TCIA) \cite{TCIA}, and specifically from the TCGA-GBM \cite{scarpace2016radiology} and the TCGA-LGG \cite{pedano2016radiology} collections. Both pre- and post-operative scans were included from these collections, and the ground truth segmentations were annotated by the fusion of previous algorithms that ranked highly during BraTS'12 and '13. During the BraTS’17 challenge all the data were revised by board-certified neuroradiologists, who assessed the complete TCIA collections (TCGA-GBM, n=262 and TCGA-LGG, n=199) and categorized each scan as pre- or post-operative, and only the scans without any prior instrumentation were included as a part of the BraTS challenge this year onwards \cite{bakas2017segmentation_1,bakas2017segmentation_2,bakas2017advancing}. In BraTS'17-'20’ the challenge was extended to the prediction of patient overall survival for the glioblastoma cases that underwent gross-total resection. This year, the BraTS 2021 challenge continues its focus on the segmentation of glioma sub-regions, with a substantially larger dataset (2,000 glioma cases = 8,000 mpMRI scans), and extends to the clinically relevant task of identifying the tumor’s MGMT promoter methylation status (methylated/unmethylated). These additional exams were obtained as a collection of the pre-operative cases of the TCIA public collections of TCGA-GBM, TCGA-LGG, IvyGAP \cite{ivygap1_puchalski2018anatomic,ivygap2_shah2016data}, CPTAC-GBM \cite{CPTAC_GBM, wang2021proteogenomic}, and  ACRIN-FMISO-Brain (ACRIN 6684) \cite{ACRIN_FMISO1, ACRIN_FMISO2}, as well as contributions from private institutional collections. The name mapping between the previous and the current challenge, as well as all the TCIA collections will be provided to further facilitate research beyond the directly BraTS related tasks.

    \subsubsection{ Tumor Annotation Protocol}
        We designed the following tumor annotation protocol, to ensure consistency in the ground truth delineations across various annotators. For the tasks related to BraTS, only structural mpMRI volumes were considered (T1, T1Gd, T2, T2-FLAIR), all of them co-registered to a common anatomical template (SRI24 \cite{SRI_rohlfing2010sri24}) and resampled to 1mm$^3$. The end to end pipeline is available for these through CaPTk \cite{captk,captk_2,captk_3} and FeTS tools. We note that radiologic definition of tumor boundaries, especially in such infiltrative tumors as gliomas, is a well-known problem. In an attempt to offer a standardized approach to assess and evaluate various tumor sub-regions, the BraTS initiative, after consultation with internationally recognized expert neuroradiologists, defined the various tumor sub-regions. However, we note that other criteria for delineation could be set, resulting in slightly different tumor sub-regions. For the BraTS 2021 challenge the regions considered are: i) the ``enhancing tumor'' (ET), ii) the ``tumor core'' (TC)  and iii) the complete tumor extent also referred to as the ``whole tumor'' (WT). The ET is described by areas that show hyper-intensity in T1Gd when compared to T1, but also when compared to ``healthy'' white matter in T1Gd. The TC describes the bulk of the tumor, which is what is typically considered for surgical excision. The TC entails the ET, as well as the necrotic (NCR) parts of the tumor, the appearance of which is typically hypo-intense in T1Gd when compared to T1. The WT describes the complete extent of the disease, as it entails the TC and the peritumoral edematous/invaded tissue (ED), which is typically depicted by the abnormal hyper-intense signal in the T2-FLAIR volume.
        
        The BraTS tumor sub-regions (visual features) are image-based and do not reflect strict biologic entities. For example, the ET regions may be defined as hyper-intense signal on T1Gd images. However, in high grade tumors, non-necrotic, non-cystic regions are present that do not enhance and can be separable from the surrounding vasogenic edema, representing non-enhancing infiltrative tumor. Another issue is defining the tumor center in low grade gliomas as it is difficult to differentiate tumor from vasogenic edema, particularly in the absence of enhancement. In the previous BraTS challenges annotators would start from the manual delineation of the abnormal signal in the T2-weighted images, primarily defining the WT, then address the TC, and finally the enhancing and non-enhancing/necrotic core, possibly using semi-automatic tools.
 
        To facilitate the annotation process for BraTS 2021, initial automated segmentations were generated by fusing previously top-performing BraTS methods. The specific methods fused were the DeepMedic \cite{kamnitsas2017efficient}, DeepScan \cite{mckinley2018ensembles}, and nnU-Net \cite{isensee2020nnu}, all trained on the BraTS 2020 dataset \cite{menze2014multimodal,bakas2017advancing,  bakas2018identifying}. The STAPLE label fusion \cite{warfield2004simultaneous} was used to aggregate the segmentation produced by each of the individual methods, and account for systematic errors generated by each of them separately. All these segmentation methods and the exact pipeline used to generate the fused automated segmentation has been made publicly available through the Federated Tumor Segmentation (FeTS) platform\footnote{\url{https://www.med.upenn.edu/cbica/fets/}} \cite{sheller2020federated}.  
         
        The volunteer neuroradiology expert annotators were provided with four mpMRI scans along with the fused automated segmentation volume to initiate the manual refinements. The ITK-SNAP \cite{itksnap} software was used for making these refinements. Once the automated segmentations were refined by the annotators, two senior attending board-certified neuroradiologists with more than 15 years of experience each, reviewed the segmentations. Depending upon correctness, these segmentations were either approved or returned to the individual annotator for further refinements. This process was followed iteratively until the approvers found the refined tumor sub-region segmentations acceptable for public release and the challenge conduction.

        \subsubsection{Common errors of automated segmentations}
\begin{figure}
        \begin{subfigure}{\textwidth}
          \centering
          \includegraphics[width=.8\linewidth]{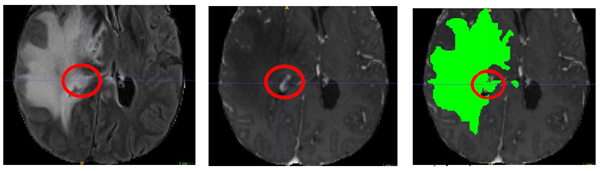}  
          \caption{Choroid plexus erroneously marked as ED.}
          \label{a}
        \end{subfigure}
        \newline
        
        \begin{subfigure}{\textwidth}
          \centering
          \includegraphics[width=.8\linewidth]{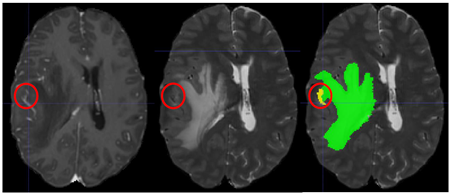}  
          \caption{Vessels in ED marked at ET.}
          \label{b}
        \end{subfigure}
        \newline
        
        \begin{subfigure}{\textwidth}
          \centering
          \includegraphics[width=.8\linewidth]{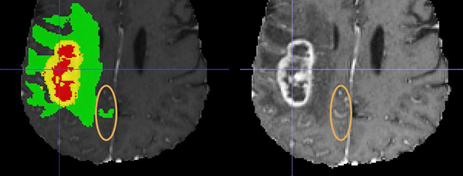}  
          \caption{Vessels in ED}
          \label{c}
        \end{subfigure}
        \newline
        
        \begin{subfigure}{\textwidth}
          \centering
          \includegraphics[width=.8\linewidth]{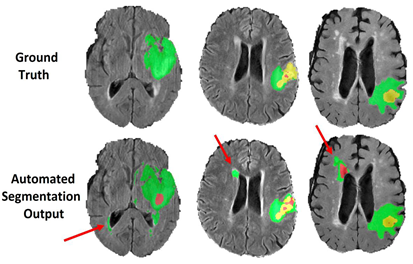}  
          \caption{Periventricular white matter hyperintensities. Figure taken from \cite{10.3389/fncom.2019.00084}.}
          \label{fig:d}
        \end{subfigure}
        \caption{Common errors expected from the automatic segmentations. }
        \label{fig:fig}
        \end{figure}

            Building upon observations during all previous BraTS instances, we note some common errors in the automated segmentations. The most typical such errors observed are:
            
            \begin{enumerate}
                \item  The choroid plexus and areas of T1 bright blood products (when they can be discriminated by comparing with the pre contrast T1 images), have erroneously been labelled as ED (Fig. \ref{a}).
                \item Vessels within the peritumoral T2 FLAIR edematous area, have been marked as ET (Fig. \ref{b}).
                \item Vessels within the peritumoral T2 FLAIR edematous area, have been marked as ED (Fig. \ref{c}).
                \item Periventricular white matter hyperintensities being confused and segmented as tumor/peritumoral regions (Fig. \ref{fig:d}).
            \end{enumerate}
     
    \subsection{Challenge Tasks}  
     
        The BraTS 2021 challenge utilizes multi-institutional mpMRI scans, and focuses on (Task 1) the evaluation of state-of-the-art methods for the segmentation of intrinsically heterogeneous brain glioblastoma sub-regions in mpMRI scans. Furthermore, to pinpoint the clinical relevance of this segmentation task, BraTS 2021 also focuses on (Task 2) the evaluation of methods to predict the MGMT promoter methylation status at the pre-operative baseline scans, via integrative analyses of quantitative imaging phenomic features and machine learning algorithms. Participants are free to choose whether they want to focus only on one or both tasks.

        \subsubsection{Task 1: Brain Tumor Sub-region Segmentation }
            The participants are called to address this task by using the provided clinically-acquired training data to develop their method and produce segmentation labels of the glioma sub-regions. The sub-regions considered for evaluation are the ``enhancing tumor" (ET), the ``tumor core" (TC), and the ``whole tumor" (WT). 
            The provided segmentation labels have values of 1 for NCR, 2 for ED, 4 for ET, and 0 for everything else. For this task this year’s BraTS challenge makes available a dataset of 8,000 MRI scans from 2,000 glioma patients. These cases are distributed across training, validation, and testing datasets following a machine learning paradigm.  
        
        \subsubsection{Task 2: Radiogenomic Classification}
        
            Participants are provided with mpMRI data and the MGMT promoter methylation status associated with each case. The methylated cases are marked as ‘1’ and unmethylated as ‘0’ in the csv file which is provided with the data. Researchers have proposed methods to predict the MGMT promoter methylation status with  appropriate imaging/radiomic features extraction, and analyse them through machine learning algorithms. The participants do not need to be limited to volumetric parameters, but can also consider intensity, morphologic, histogram-based, and textural features, as well as spatial information, and glioma diffusion properties extracted from glioma growth models. Participants will be evaluated for the predicted MGMT status of the subjects indicated in the accompanying spreadsheet.
            
    \subsection{Performance Evaluation}  
    
        Participants are called to submit the results on the online evaluation platform for the training and validation dataset. The test dataset will never be shared with the participants and they will upload their proposed methods in a containerized way for the final testing phase.  To evaluate the generalizability of the proposed methods, we will evaluate the performance on the cohort which is not part of either training or validation cohort, also termed as testing out of distribution cohort. The distribution for methylated and unmethylated cases across the training, validation, testing cohort is given in Table \ref{Table1}.  

        \subsubsection{Task 1: Tumor Sub-region Segmentation}
        Consistent with the configuration of previous BraTS challenges, we intend to use the ``Dice similarity coefficient", and the ``Hausdorff distance (95\%)" as performance evaluation metrics. Expanding upon this evaluation scheme, we will also provide the metrics of ``Sensitivity" and ``Specificity", allowing to determine potential over- or under-segmentations of the tumor sub-regions by participating methods. 
 
        The ranking scheme followed during the BraTS 2017-2020 comprised the ranking of each team relative to its competitors for each of the testing subjects, for each evaluated region (i.e., ET, TC, WT), and for each measure (i.e., Dice and Hausdorff). For example, in BraTS 2020, each team was ranked for 166 subjects, for 3 regions, and for 2 metrics, which resulted in $166\times3\times2=996$ individual rankings. The final ranking score (FRS) for each team was then calculated by firstly averaging across all these individual rankings for each patient (i.e., Cumulative Rank), and then averaging these cumulative ranks across all patients for each participating team. This ranking scheme has also been adopted in other challenges with satisfactory results, such as the Ischemic Stroke Lesion Segmentation challenge\footnote{\url{http://www.isles-challenge.org/}}  \cite{maier2017isles}.
        
        We then conducted further permutation testing, to determine statistical significance of the relative rankings between each pair of teams. This permutation testing would reflect differences in performance that exceeded those that might be expected by chance. Specifically, for each team we started with a list of observed subject-level Cumulative Ranks, i.e., the actual ranking described above. For each pair of teams, we repeatedly randomly permuted (i.e., for 100,000 times) the Cumulative Ranks for each subject. For each permutation, we calculated the difference in the FRS between this pair of teams. The proportion of times the difference in FRS calculated using randomly permuted data exceeded the observed difference in FRS (i.e., using the actual data) indicated the statistical significance of their relative rankings as a p-value. These values were reported in an upper triangular matrix providing insights of statistically significant differences across each pair of participated teams.
         
        Top ranked methods in the validation phase will be invited at MICCAI 2021 for presentation of their methods and results. The final top three ranked participating teams according to their evaluation against the testing data, will be invited at RSNA 2021 for presentation and to receive their monetary awards.

         \subsubsection{Task 2: Radiogenomic Classification}
         The methods submitted by the participating teams for task 2 will be evaluated based on the area under the ROC curve (AUC), accuracy, FScore (Beta) and Matthew's Correlation Coefficient of the classification of the MGMT status as methylated and unmethylated. The AUC is a metric that measures the overall discriminatory capacity of a model for all possible thresholds and allows for comparing the performance of the entries by each participant, even though it has no straightforward clinical meaning and does not guarantee the model is calibrated. The AUC will be used as the reference metric to rank the participants in the leaderboard of task 2.

        \subsection{Participation Timeline}  
        
        The challenge will commence with the release of the training dataset, which will consist of imaging data and the corresponding ground-truth labels. Participants can start designing and training their methods using this training dataset.
     
        The validation data will then be released within three weeks after the training data is released. This will allow participants to obtain preliminary results in unseen data and also report these in their submitted short MICCAI LNCS papers, in addition to their cross-validated results on the training data. The ground truth of the validation data will not be provided to the participants, but multiple submissions to the online evaluation platforms will be allowed. The top-ranked participating teams in the validation phase will be invited to prepare their slides for a short oral presentation of their method during the BraTS challenge at MICCAI 2021. 
         
        Finally, all participants will be evaluated and ranked on the same unseen testing data, which will not be made available to the participants, after uploading their containerized method in the evaluation platforms. The final top-ranked participating teams will be announced at the 2021 RSNA Annual Meeting. The top-ranked participating teams of both the tasks will receive monetary prizes of total value of \$60,000, sponsored by Intel, RSNA, and NeoSoma Inc.

\section{Discussion}

In this paper we presented the design of the $10^{th}$ BraTS challenge, jointly organised by the RSNA, ASNR, and MICCAI societies, and offering what can possibly be considered the largest curated multi-label annotated dataset of mpMRI scans for a single disease. Members of the RSNA and ASNR communities had graciously volunteered to refine tumor sub-region annotations for all 2,000 cases included in the BraTS 2021 challenge, until satisfactory quality for releasing the data. Considering the size of this year’s challenge and also its potential continuation after the announcement of this year’s winners, the testing data will be kept hidden at all times and their performance evaluation will be based on the challenge evaluation platforms of Sage Bionetworks Synapse (Task 1) and Kaggle (Task 2), concluding in distributing to the top ranked participants monetary awards of \$60,000 collectively. We hope that the well-labelled multi-institutional data of BraTS 2021 will provide an optimized community benchmark and a common dataset to the research community focusing on computational neuro-oncology, even beyond the specific BraTS 2021 tasks.

Although we designed the BraTS 2021 challenge with utmost care there are still some limitations that need further consideration. Firstly, the tumor feature segmentations of each case are refined by a single annotator with an iterative process with a group of approvers, until approval from the latter, and hence the potential inter-rater agreement can not be assessed. Secondly, since the provided MGMT promoter methylation status was determined based on varying methods across the multiple institutions that contributed data, and each institute follows its own methodology (e.g., pyrosequencing vs quantitative PCR) and thresholds, only a binary classification of the methylation status was made available to the participants instead of a continuous value. Lastly, we note that some of the MRI datas included in the challenge harbor more abnormalities than just gliomas. Since the focus of the challenge was on gliomas all other abnormalities (such as white matter hyperintensities that are typically secondary to small vessel ischemic disease) were not considered in the annotation process. This was made particularly apparent from previous efforts that attempted to perform a multi-disease segmentation \cite{10.3389/fncom.2019.00084}. 

With this multi-disease segmentation in mind, one of the main future directions for the BraTS challenge would be to expand beyond its current focus on glial tumors towards general brain abnormalities. Furthermore, the extension from solely pre-operative baseline scans to post-operative scans, and the inclusion of an additional label for the resection cavity would be a very interesting and clinically appealing direction, as it would speak directly to the assessment of treatment response and disease progression. To ensure robustness and generalizability of the computational algorithms, ample patient data from multiple sites, capturing diverse patient populations are desired. A major hindrance for accessing these datasets is data siloing due to tedious bureaucratic process, data ownership concerns, and legal considerations reflected in patient privacy regulations, such as the American HIPAA \cite{hippa} and the European GDPR\cite{gdpr}. In future, we aim at moving from the current centralised data approach to a federated approach, which would enable researchers to access potentially unprecedented size of data and hence design more robust and generalizable algorithms \cite{sheller2020federated, rieke2020future, pati2021federated, pati2021gandlf}.

\section*{Acknowledgments}
    Success of any challenge in the medical domain depends upon the quality of well annotated multi-institutional datasets. We are grateful to all the data contributors, annotators and approvers for their time and efforts.

\section*{Funding}

Research reported in this publication was partly supported by the National Cancer Institute (NCI) Informatics Technology for Cancer Research (ITCR) program and the National Institute of Neurological Disorders and Stroke (NINDS) of the National Institutes of Health (NIH), under award numbers NCI:U01CA242871, NCI:U24CA189523, NINDS:R01NS042645, Contract No. HHSN261200800001E, Ruth L. Kirschstein Institutional National Research Service Award number T32 EB001631. Research reported in this publication was also partly supported by the RSNA Research \& Education Foundation grant number RR2011, and by the ASNR Foundation Grant in Artificial Intelligence (JDR). Sage Bionetworks support of challenge organization and infrastructure was supported by the NCI ITCR program under award number U24CA248265. The content of this publication is solely the responsibility of the authors and does not represent the official views of the NIH or of the RSNA R\&E Foundation, or the views or policies of the Department of Health and Human Services, nor does mention of trade names, commercial products, or organizations imply endorsement by the U.S. Government.

\bibliographystyle{ieeetr}
\bibliography{bibliography.bib}

\end{document}